\documentclass[a4paper, 10pt, conference]{mhs}      

\IEEEoverridecommandlockouts                              

\usepackage{graphics} 
\usepackage{epsfig} 
\usepackage{mathptmx} 
\usepackage{times} 
\usepackage{amsmath} 
\usepackage{amssymb}  
\usepackage{here}
\usepackage{threeparttable}
\usepackage{caption}

\title{\Large \bf
Embedded Image-to-Image Translation for Efficient Sim-to-Real Transfer in Learning-based Robot-Assisted Soft Manipulation}

\author{{\large Jacinto Colan$^{1}$, Keisuke Sugita$^{1}$, Ana Davila$^{2}$, Yutaro Yamada$^{1}$ and Yasuhisa Hasegawa$^{2}$}\\
{\normalsize $^{1}$ Dept. of Micro-Nano Mechanical Science and Engineering, Nagoya University, Aichi, Nagoya, Japan}\\
{\normalsize $^{2}$ Institutes of Innovation for Future Society, Nagoya University, Aichi, Nagoya, Japan}\\\\}

\begin{document}

\maketitle
\thispagestyle{empty}
\pagestyle{empty}

\begin{abstract}
    Recent advances in robotic learning in simulation have shown impressive results in accelerating learning complex manipulation skills. However, the sim-to-real gap, caused by discrepancies between simulation and reality, poses significant challenges for the effective deployment of autonomous surgical systems. We propose a novel approach utilizing image translation models to mitigate domain mismatches and facilitate efficient robot skill learning in a simulated environment. Our method involves the use of contrastive unpaired Image-to-image translation, allowing for the acquisition of embedded representations from these transformed images. Subsequently, these embeddings are used to improve the efficiency of training surgical manipulation models. We conducted experiments to evaluate the performance of our approach, demonstrating that it significantly enhances task success rates and reduces the steps required for task completion compared to traditional methods. The results indicate that our proposed system effectively bridges the sim-to-real gap, providing a robust framework for advancing the autonomy of surgical robots in minimally invasive procedures.
\end{abstract}


\section{Introduction}

Laparoscopic surgery, a minimally invasive technique, has gained widespread adoption due to its numerous benefits, including reduced postoperative pain, faster recovery times, and minimal scarring. Despite these advantages, laparoscopic surgery presents significant challenges, primarily due to the lack of direct tactile feedback \cite{colan2022review} and the reliance on visual information from laparoscopic cameras \cite{fozilov2023endoscope, yamada2024multimodal}. This complexity underscores the need for advanced robotic assistance to improve surgical precision and efficiency.

Simulation-based learning has emerged as a promising approach to train autonomous surgical robots in a controlled, risk-free environment. This method allows for the optimization of AI models using virtual environments, thus eliminating the need for early-stage deployment in real-world scenarios, which can be risky and ethically challenging. Successful implementations for surgical applications have been demonstrated in previous work for robotic manipulation learning \cite{xu2021surrol, scheikl2023lapgym, yu2024orbit}. However, a significant obstacle remains: the sim-to-real gap. This gap refers to performance discrepancies that arise when models trained in simulation are applied to real-world scenarios, often due to differences in physical properties and visual characteristics between simulated and real environments.

Addressing the sim-to-real gap is crucial for the effective deployment of autonomous surgical systems in clinical settings. Previous research has explored various methods to bridge this gap, including domain randomization and domain adaptation techniques. These methods involve varying the parameters of the simulation to expose the model to a wide range of scenarios, which helps to generalize to real-world conditions. However, this method can be computationally intensive and may not fully capture the complexities of real-world surgical environments \cite{ou2023simtoreal}.

In this paper, we propose a novel approach to mitigate the sim-to-real gap in laparoscopic surgery by employing image translation models. Our method involves training an image translation model to convert simulated images into realistic counterparts, creating a more seamless transition from simulation to reality. By obtaining embedded representations from these transformed images, we enhance the training efficiency and accuracy of surgical manipulation models, particularly when dealing with high-resolution images.

\begin{figure}[t]
    \centering
    \includegraphics[width=0.9\columnwidth]{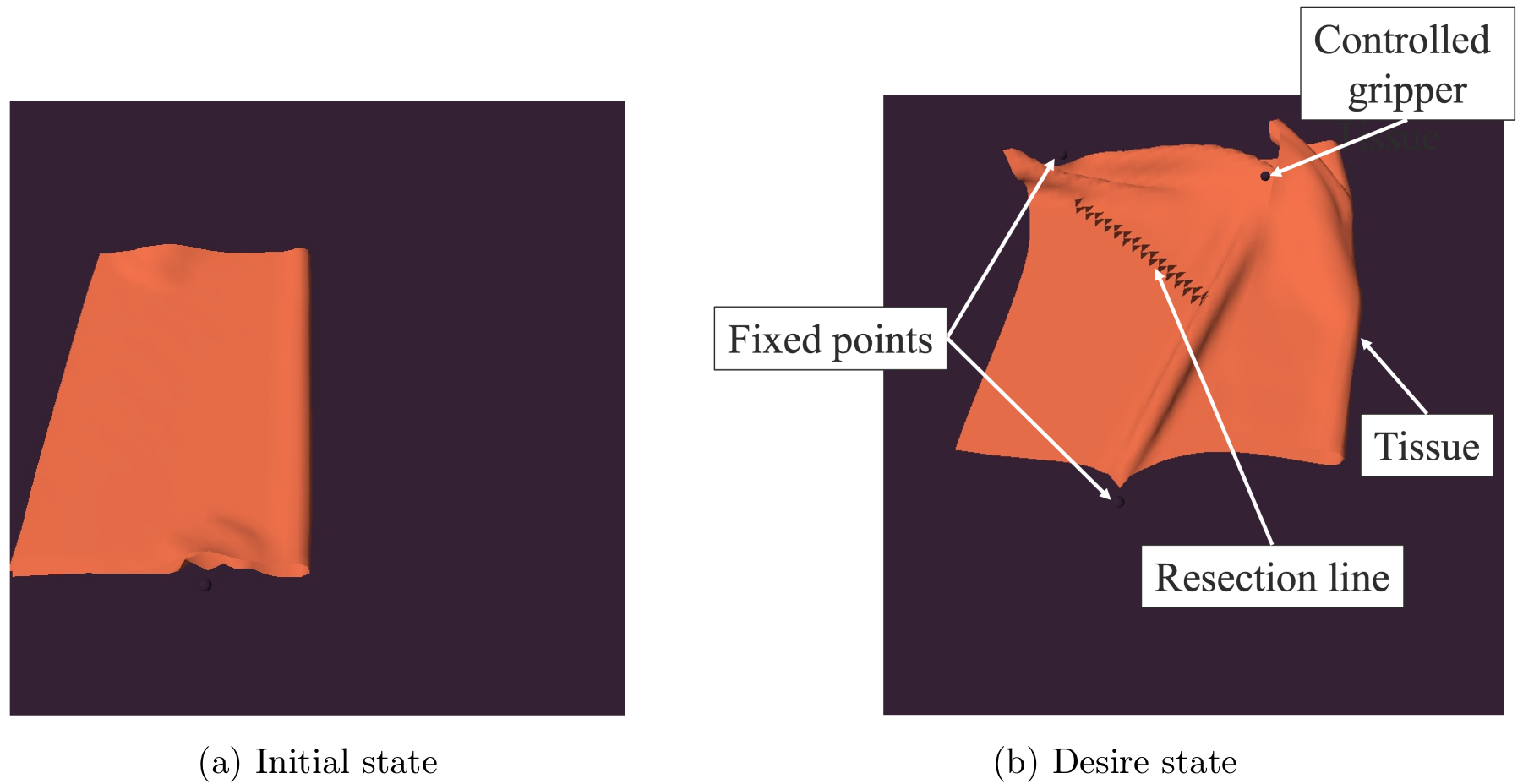}
    \caption{Simulation of a dummy tissue used for learning-based tissue triangulation, aimed at visualizing the resection path.}
    \label{fig:1}
\end{figure}


\section{Methodology}
We propose the use of embedding representations for a Generative Adversarial Network (GAN)-based image translation model to accelerate the conversion between simulated images and their realistic counterparts.

\subsection{Image translation model}

Unpaired Image-to-Image Translation (UI2I) \cite{zhu2017unpaired} has proven effective in generating real-world images from simulation images. UI2I maps between different image domains using asymmetric training data, making it suitable for the dynamic and unpredictable nature of surgical environments. In this study, we focus on Contrastive Unpaired Translation (CUT) \cite{park2020contrastive}, which incorporates mutual information between images from two domains into the loss function using contrastive learning.

CUT follows a GAN framework \cite{goodfellow2014gan}, comprising a generator (G) and a discriminator (D).  The generator is divided into an encoder ($G_{\text{enc}}$) and a decoder ($G_{\text{dec}}$). Let $X$ be the source domain and $Y$ the target domain, with image datasets $X = \{x \in X\}$ and $Y = \{y \in Y\}$. The output image from the generator, $\hat{y}$, is obtained as $\hat{y} = G(z) = G_{\text{dec}}(G_{\text{enc}}(x))$

To train CUT, the loss function consists of two parts: Adversarial loss (GAN loss) and Patchwise Contrastive loss (PatchNCE loss). The GAN loss encourages the output image to visually resemble images from the target domain. In this study, we use the Least Squares GAN (LSGAN) loss function \cite{mao2017lsgan} to prevent instability due to vanishing gradients from the sigmoid function. The objective functions for LSGAN are as follows, where $a$, $b$, and $c$ are constants:

\begin{equation}
    \begin{aligned}
        \min_{D} J(D) & = \frac{1}{2} \mathbb{E}_{y \in Y} [(D(y) - b)^2] + \frac{1}{2} \mathbb{E}_{x \in X} [(D(G(x)) - a)^2], \\
                      & = \frac{1}{2} \mathbb{E}_{x \in X} [(D(G(x)) - c)^2]
    \end{aligned}
\end{equation}

PatchNCE loss aims to maximize mutual information between the input and output images, extending contrastive loss to multiple layers and patches. During the calculation of contrastive loss, query patches are sampled from the output image, and positive patches from the same location and $N$ negative patches from different locations are sampled from the input image. These patches are mapped to $K$-dimensional vectors $v \in \mathbb{R}^K$, $v^+ \in \mathbb{R}^K$, and $v^- \in \mathbb{R}^{N \times K}$, respectively, and normalized. Next, the cosine similarity between the query patch and both positive and negative patches is calculated and scaled by a temperature parameter $\tau$. These are treated as logits, and cross-entropy is computed to represent the probability of selecting the positive patch in a $(N+1)$ class classification problem. The NCE loss is obtained by multiplying this probability by -1. The NCE loss, calculated from $v$, $v^+$, and $v^-$, is given by:

\begin{equation}
    \ell(v, v^+, v^-) = -\log \left( \frac{\exp(v \cdot v^+ / \tau)}{\exp(v \cdot v^+ / \tau) + \sum_{n=1}^N \exp(v \cdot v^- / \tau)} \right)
\end{equation}

For the calculation of PatchNCE loss, $L$ layers from the feature extractor are selected, and $S_l$ query patches are sampled from the output of the $l$-th layer. Contrastive loss is calculated for all these patches. Let $l \in \{1, 2, \ldots, L\}$ and $s \in \{1, 2, \ldots, S_l\}$, then the PatchNCE loss is given by:

\begin{equation}
    L_{\text{PatchNCE}}(X) = \mathbb{E}_{x \sim X} \left[ \sum_{l=1}^L \sum_{s=1}^{S_l} \ell(\hat{z}_l^s, z_l^s, z_l) \right]
\end{equation}

where $z$ and $\hat{z}$ are the features calculated from the input image $x$ and the output image $\hat{y}$, respectively.

The overall loss $L_{\text{CUT}}$ in CUT is computed as

\begin{equation}
    L_{\text{CUT}} = \lambda_{\text{GAN}} L_{\text{GAN}}(X, Y) + \lambda_X L_{\text{PatchNCE}}(X) + \lambda_Y L_{\text{PatchNCE}}(Y)
\end{equation}

where $L_{\text{GAN}}(X, Y)$ is the GAN loss for the generator as described above, and $\lambda_{\text{GAN}}$, $\lambda_X$, $\lambda_Y$ are weighting parameters. $L_{\text{PatchNCE}}(Y)$ is added to prevent unnecessary changes by the generator to the target domain images:

\subsection{Embedded representations using the image translation model}
During CUT training, the encoder part of the generator ($G_{\text{enc}}$) and the projection head ($H$), consisting of several layers of MLP, are used as the feature extractor for loss calculation. In our proposed system, this feature extractor is also used during the training of the tissue manipulation model, allowing training with the embedded representations of images as inputs.

We select $L$ layers from $G_{\text{enc}}$, and the output from the $l$-th selected layer is denoted as $G_{\text{enc}}^l$. During feature extraction, the output image $\hat{y}$ is input into $G_{\text{enc}}$, and the outputs from each selected layer are calculated. From these outputs, $S_l$ patches are selected and input into the projection head $H$, mapping them to a $k$-dimensional vector space. Let $\hat{z}^l$ be the features obtained from the $l$-th layer of $G_{\text{enc}}$:

\begin{equation}
    \hat{z}^l = H(G_{\text{enc}}^l(\hat{y}))
\end{equation}

The combined features from all layers, denoted $\{\hat{z}^l\}_L$, form a vector of dimensions $\{\hat{z}^l\}_L \in \mathbb{R}^{L \times S \times k}$. By adjusting $L$, $S$, and $k$, the input data size can be modified. During training of the tissue manipulation model, these features are flattened and input into fully connected layers, enabling training with the same input data size even for high-resolution images. This can significantly reduce the data size for training \cite{park2020contrastive}. Additionally, since there is no need to update the feature extractor for image input during the training of the tissue manipulation model, the proposed system can potentially improve learning efficiency.

One of the reasons for adopting CUT as the training algorithm for the image translation model in this study is that CUT uses a common encoder and projection head for both the source and target domains. In contrast, DCL (Dual Contrastive Learning) \cite{han2021dual}, an improved version of CUT and another UI2I model training algorithm, uses separate feature extractors for each domain, leading to different feature extractors for real-world and simulation images. CUT, on the other hand, allows the use of a common feature extractor for both simulation and real-world images, enabling the acquisition of task-agnostic features regardless of image domain differences.


\section{Implementation}

\subsection{Simulation environment}
Our simulation environment is based on Softgym \cite{lin2021softgym}, which uses NVIDIA FleX to enable seamless interaction of rigid bodies, fluids, and deformable objects using a unified particle representation. The environment models tissues as deformable objects and surgical tools as rigid bodies, facilitating realistic interactions. We used PyFleX \cite{li2018learning}, a Python binding of NVIDIA FleX, to integrate the simulation with our machine learning algorithms.

\subsection{Task characterization for tissue manipulation learning}
A triangulation task \cite{wang2012robotics}, frequently performed in Robot-Assisted Minimally Invasive Surgery (RAMIS), is set as the learning task. Triangulation, as shown in Figure~\ref{fig:1}, involves pulling the tissue into a triangular shape using three forceps \cite{liu2024latent}. This technique is used primarily during tissue resection to make the tissue tense and easier to cut. The soft tissue is modeled as a rectangle measuring 8 cm × 10 cm, with two points on the edge fixed by non-controlled grippers. The background color is set similar to the real environment for training the image translation model. The tissue is initially placed at the position $[0.335, 0.102, 0.465]$ m. Based on discussions with surgeons experienced in da Vinci surgery, the objectives of the triangulation task are defined as follows:

\begin{itemize}
    \item Objective 1: Visualization of the resection area.
    \item Objective 2: Positioning the resection area within the working range.
\end{itemize}

The resection line is represented by a straight line with the initial state set where the resection line is hidden, as shown in Figure~\ref{fig:1}. The triangulation working area is limited to the interior of the triangle formed by the tips of the three forceps, so Objective 2 is equivalent to positioning the resection line within this triangle.

\subsubsection{Action space}
The actions output by the model are described by:

\begin{equation}
    s_t = [p_t, d_t]
\end{equation}

where $p_t = [x_p, y_p, z_p] \in \mathbb{R}^3$ represents the tissue grasping point position, and $d_t = [x_d, y_d, z_d] \in \mathbb{R}^3$ represents the target gripper position after grasping (i.e. pulling the tissue). The output at each time step represents a sequence of operations between grasping and pulling the tissue, repeated to perform the task.

\begin{figure}[t]
    \centering
    \includegraphics[width=\columnwidth]{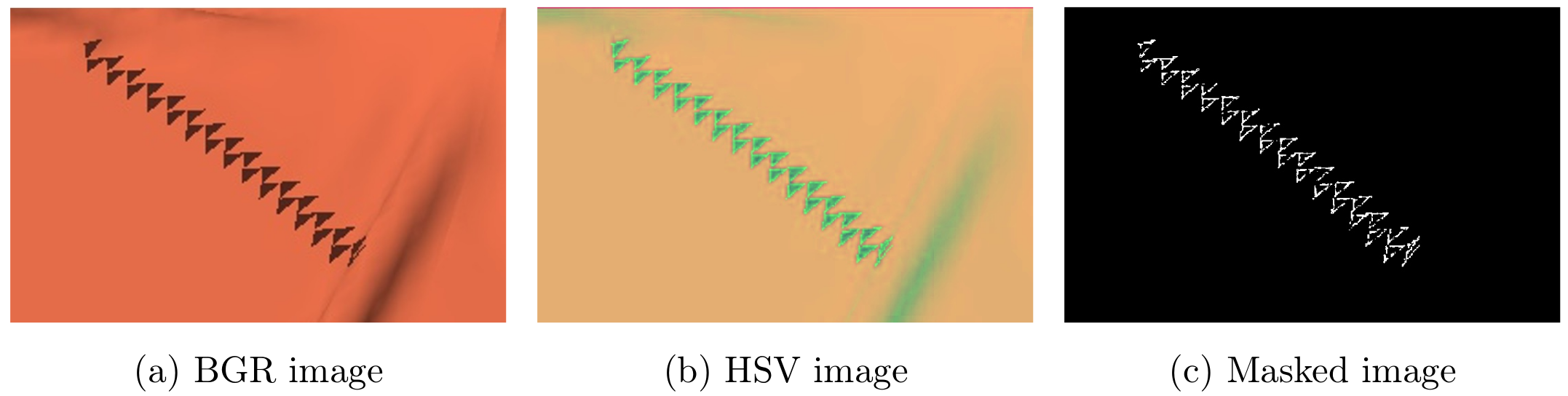}
    \caption{Image preprocessing for resection path recognition.}
    \label{fig:2}
\end{figure}

\subsubsection{Reward Function}
The reward function used to evaluate the model's performance is designed to reflect the achievement of the objectives for the triangulation task. For Objective 1, image preprocessing is performed using the OpenCV image processing library as shown in Figure~\ref{fig:2}. For reward calculation at each time step, the image obtained from the simulator's camera is extracted in BGR format and converted to HSV format. For color extraction, upper and lower bounds are set for each HSV element, and only pixels within these ranges are extracted, isolating the color of the resection line from the simulation image. The extracted result is shown as a white region in Figure~\ref{fig:4}c. Finally, the total number of extracted pixels is counted and if this number exceeds a threshold, Objective 1 is considered to be achieved.

For Objective 2, the condition that the resection line is within the triangle formed by the three grippers is equivalent to having both ends of the resection line within the triangle. Thus, Objective 2 is considered achieved if both ends of the resection line meet the following conditions:

\begin{itemize}
    \item \textbf{Condition (i)}: The endpoint projected onto the plane of the triangle formed by the grippers lies within the triangle.
    \item \textbf{Condition (ii)}: The distance between the endpoints of the resection line and their projection over the plane of the triangle are below a threshold $\epsilon_2$.
\end{itemize}

\begin{figure}[t]
    \centering
    \includegraphics[width=0.6\columnwidth]{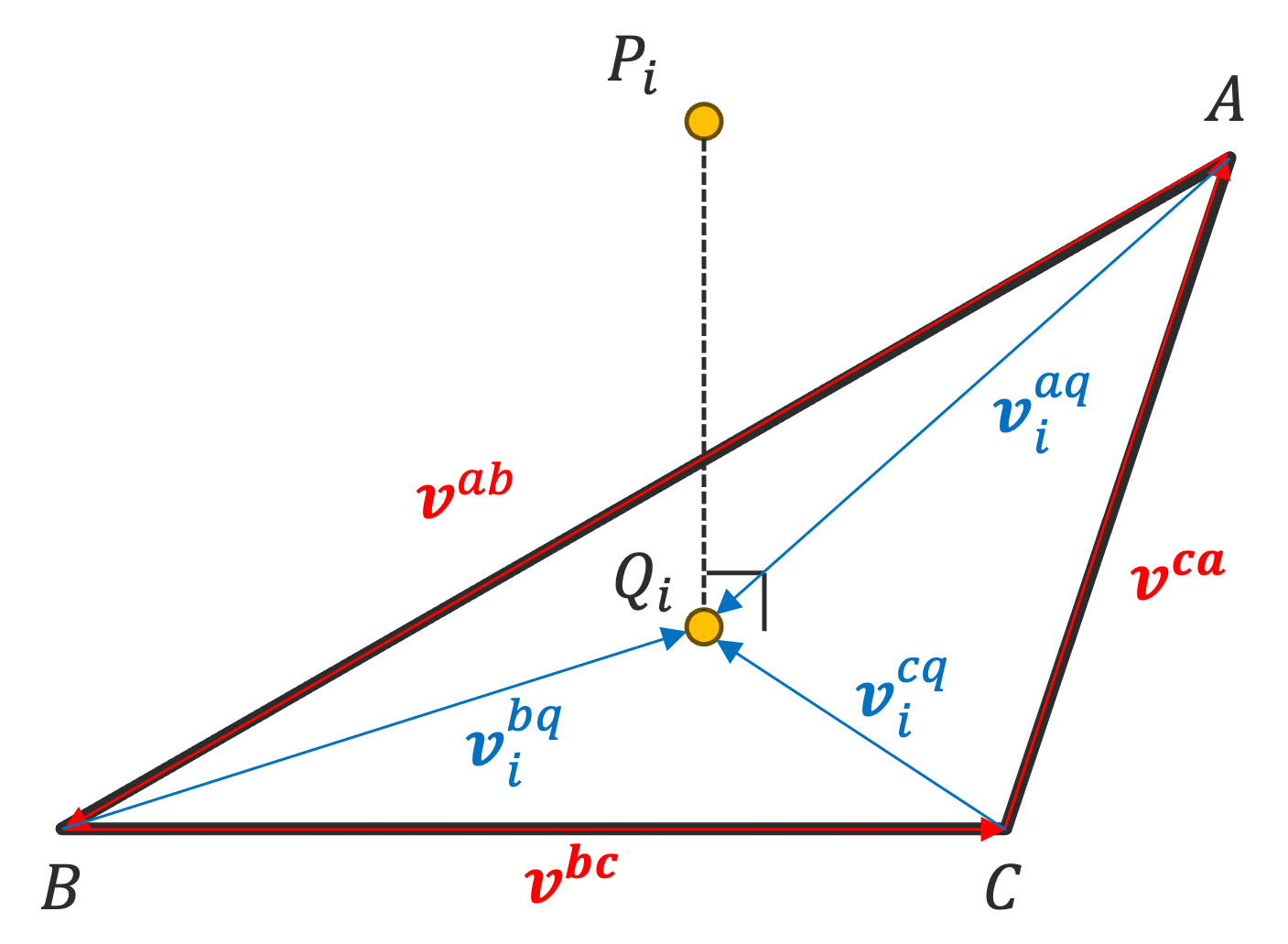}
    \caption{Verification of resection line is located inside the triangulation area. $P_i$ represents the endpoints $i={1,2}$ of the resection line, $A$, $B$, and $C$ represent the positions of the grippers, and $Q_i$ represents the points projected onto the plane defined by $A$, $B$, and $C$.}
    \label{fig:3}
\end{figure}

Figure~\ref{fig:3} shows the variables used for verification of each condition, from which the determination of $Q_i$ inside the triangle is given by first computing the internal variables:
\begin{equation}
    \begin{aligned}
        v_{i}^1 & = v^{ab} \times v_{i}^{bq} \\
        v_{i}^2 & = v^{bc} \times v_{i}^{cq} \\
        v_{i}^3 & = v^{ca} \times v_{i}^{aq}
    \end{aligned}
\end{equation}

When $v_{i1} \cdot v_{i2} \geq 0$ and $v_{i1} \cdot v_{i3} \geq 0$, point $Q_i$ is within triangle $ABC$, satisfying Condition (i) for point $P_i$. For Condition (ii), the distance between $P_i$ and $Q_i$ is calculated, and if it is below a threshold $\epsilon_2$, the condition is satisfied.

A reward is given according to Equation~\ref{eq:reward_function}.
\begin{equation}
    r(t) =
    \begin{cases}
        0   & \text{if goal 1 is not satisfied}  \\
        0.5 & \text{if only goal 1 is satisfied} \\
        1   & \text{if both goals are satisfied}
    \end{cases}
    \label{eq:reward_function}
\end{equation}

Objectives 1 and 2 are defined as follows, where $n_{mask}$ is the number of pixels masked by OpenCV's color extraction, and $\epsilon_1$ and $\epsilon_2$ are thresholds:
\begin{equation}
    \begin{aligned}
         & \text{goal 1} \iff n_{mask} \geq \epsilon_1,                                 \\
         & \text{goal 2} \iff \forall i \in \{1, 2\}, \text{(i)} \land \text{(ii)},     \\
         & \text{(i)} \iff v_{i1} \cdot v_{i2} \geq 0 \land v_{i1} \cdot v_{i3} \geq 0, \\
         & \text{(ii)} \iff |P_iQ_i| \leq \epsilon_2
    \end{aligned}
\end{equation}

\subsection{Embedding representations}
The embedding representations were extracted using the parameters $L = 5$, $S = 32$, and $k = 32$. The input data size calculated from these values is 5120, which is about 2.4\% of the original data size when using grayscale images of the standard resolution for laparoscopic images (2048 $\times$ 1080 = 2211840).

\subsection{Data collection}
\begin{figure}[t]
    \centering
    \includegraphics[width=0.8\columnwidth]{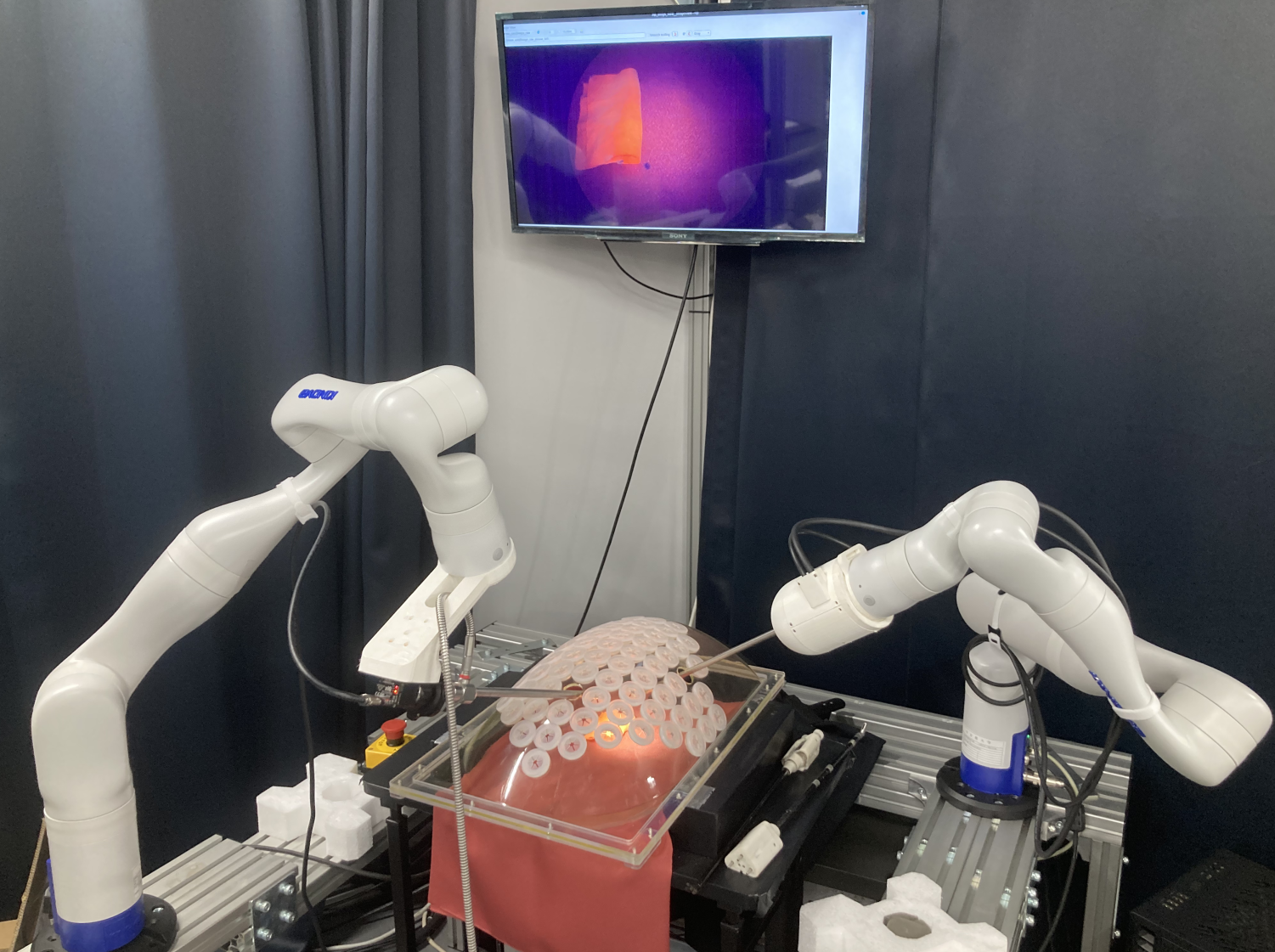}
    \caption{Experimental setup used for data collection.}
    \label{fig:4}
\end{figure}

The robotic system used for image collection in the real world comprises a robotic endoscope holder and a robotic manipulator with articulated forceps \cite{davila2024realtime}, aimed at manipulating a phantom tissue for the triangulation task. The setup is shown in Figure~\ref{fig:4}.

\subsection{Training of the Image Translation Model}
\begin{figure}[t]
    \centering
    \includegraphics[width=0.8\columnwidth]{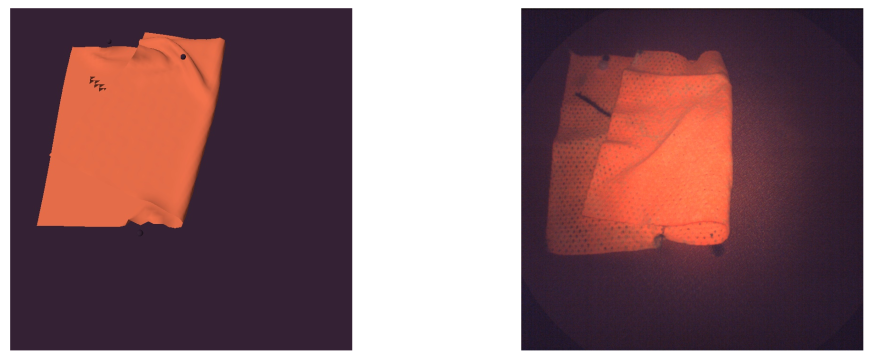}
    \caption{Examples of source and target domain images. \textbf{Left.} Dummy tissue in simulation. \textbf{Right.} Real-world dummy tissue.}
    \label{fig:5}
\end{figure}

The CUT algorithm is implemented for the image translation model. The source domain corresponds to simulator images and the target domain corresponds to real-world images, with images collected during the execution of the triangulation task in each domain. Examples of images from the source and target domains are shown in Figure~\ref{fig:5}. The image size is 512×512 for both domains, and the number of images used for training was 500 for the source domain and 150 for the target domain. The images were converted to grayscale.

An Inception-v3 architecture was used for the image translation model \cite{szegedy2016rethinking}, fine-tuned with the customized dataset \cite{davila2024comparison}. The selection of the trained image translation model was based on the Inception Score (IS) \cite{salimans2016improved} and the Frechet Inception Distance (FID) \cite{heusel2017gans}. IS evaluates the quality and diversity of generated images, while FID compares generated images with real images.
During CUT training, the GAN loss is used, where the generator and discriminator learn adversarially. The trained models are saved at regular intervals during training, and after training, IS and FID are calculated for each saved model to determine which model to use to train the tissue manipulation model. In this study, the image translation model was trained for 400 epochs, saving the model every 10 epochs. After training, IS and FID were calculated for each saved model, ranked, and the top 5 models based on the sum of their ranks in IS and FID were selected as candidates. The final selection of the model for training the tissue manipulation model involved human confirmation of the generated images.

\section{Experimental Validation}

\subsection{Model configurations for tissue manipulation learning}
This experiment aims to verify the effectiveness of the proposed system in improving the learning efficiency of the tissue manipulation model by using the following three configurations with different training inputs and comparing the results:

\begin{figure}[t]
    \centering
    \includegraphics[width=\columnwidth]{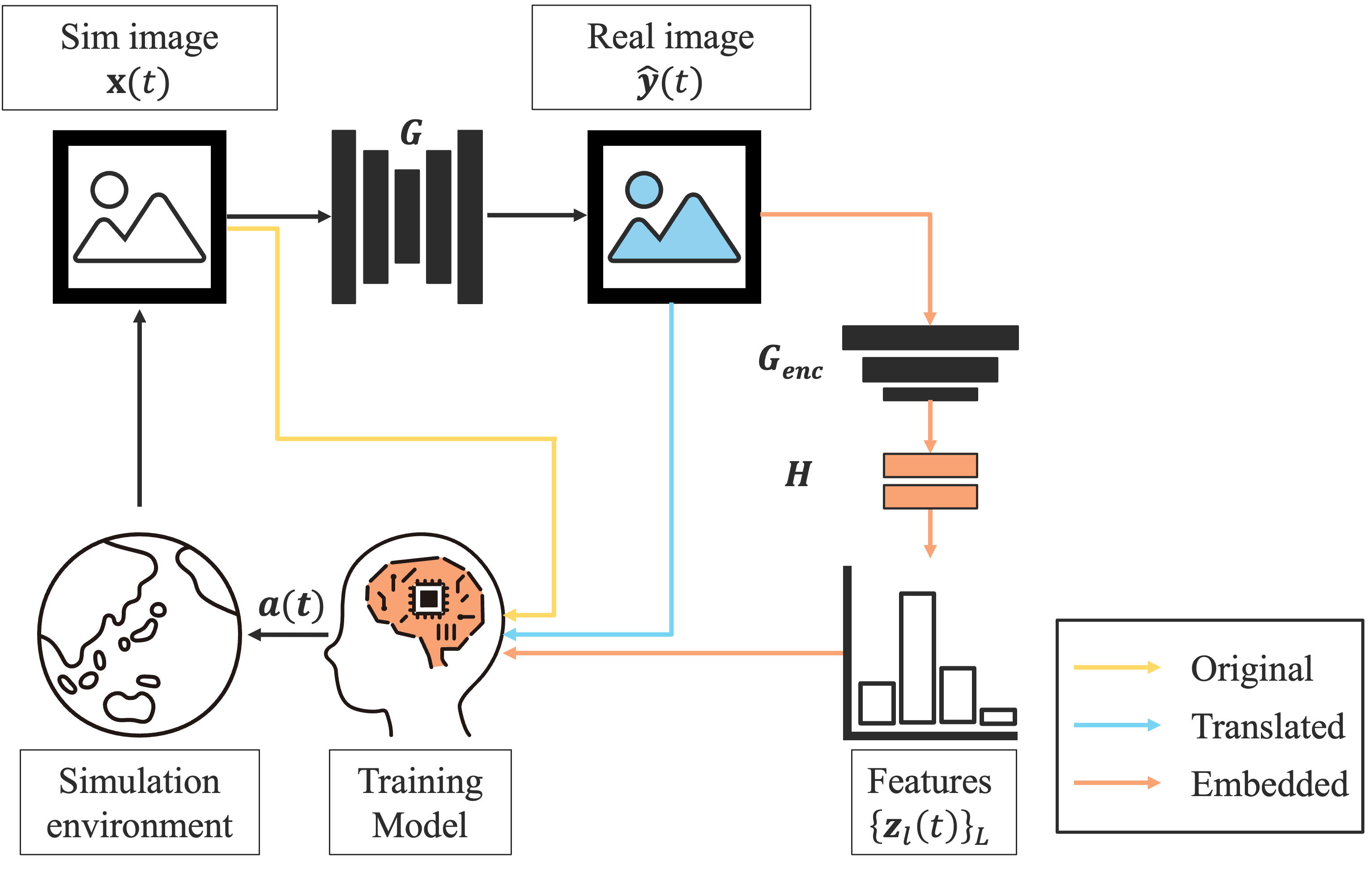}
    \caption{Diagram showing the various model configurations depending on training inputs.}
    \label{fig:6}
\end{figure}

\begin{itemize}
    \item \textbf{Original}: Simulation images.
    \item \textbf{Translated}: Real-world images obtained through the image translation model.
    \item \textbf{Embedded}: Embedded representations acquired by the proposed system.
\end{itemize}

The Original configuration uses images obtained directly from the simulator as input without using the image translation model (SBML). To match the conditions of the other two inputs, the simulation images are converted to grayscale before being fed into the model. The Translated configuration involves training using real-world images output by the image translation model, under the same conditions as in previous studies \cite{scheikl2022simtoreal}. The Embedded configuration involves training using the embedded representations acquired by the proposed system. Figure~\ref{fig:6} shows the flow of inputs to the model for each condition.

The trained models are evaluated using the following metrics:

\begin{itemize}
    \item Loss reduction speed across training iterations.
    \item Task reward across training iterations.
    \item Task success rate and the number of steps for success.
\end{itemize}

During training, the loss calculated at each policy update is recorded, and the model at each step is saved at regular intervals. After all training is completed, test episodes are conducted for all saved models, and the loss, reward, task success rate, and number of steps to success are recorded. This experiment aims to verify the improvement in learning efficiency by comparing the performance on these metrics between training iterations. Each condition is trained 10 times, saving 10 models per training by dividing the total training steps into 10. Thus, a total of 300 models are tested. The total training steps are 12800 for Original and Translated, and 128000 for Embedded. Each model undergoes 10 test episodes. The hyperparameters used for training are listed in Table~\ref{table:training_conditions}.

\begin{table}[h]
    \centering
    \caption{Training conditions for triangulation task}
    \begin{tabular}{|c|c|}
        \hline
        \textbf{Parameters}     & \textbf{Values} \\
        \hline
        Number of training data & 10              \\
        Batch size              & 64              \\
        Entropy coefficient     & 0               \\
        Learning rate           & 0.0003          \\
        Epochs                  & 128             \\
        Optimizer               & Adam            \\
        \hline
    \end{tabular}
    \label{table:training_conditions}
\end{table}

\begin{figure}[t]
    \centering
    \includegraphics[width=0.8\columnwidth]{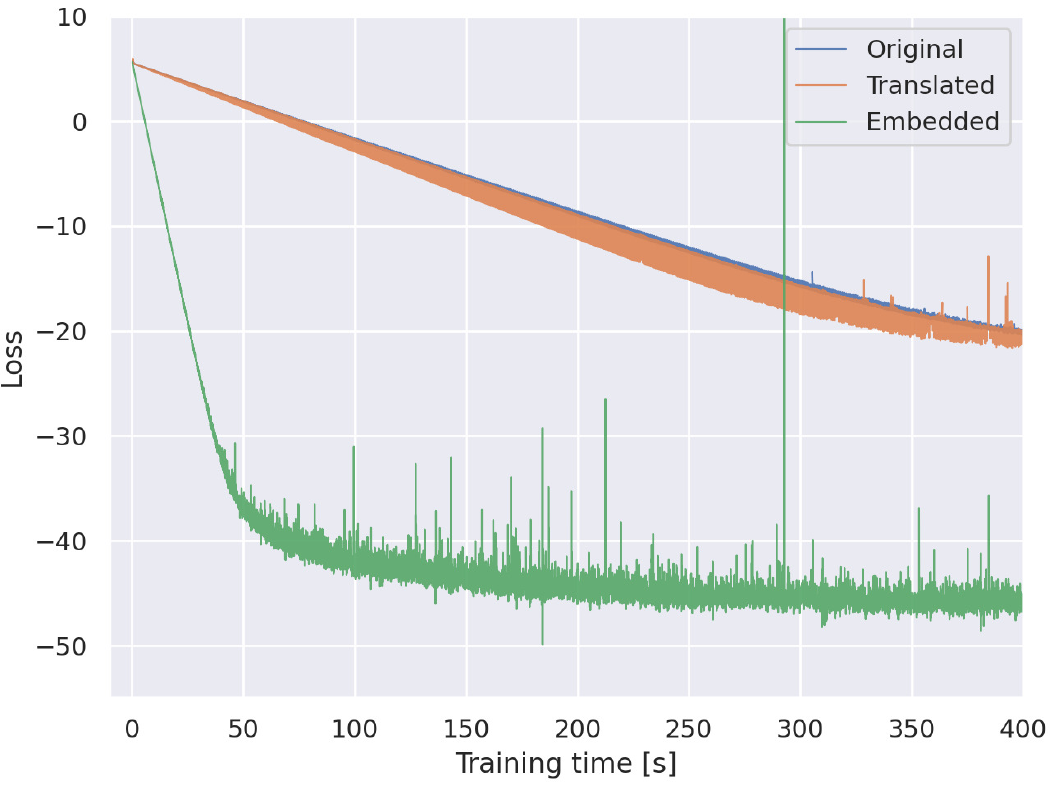}
    \caption{Training loss evolution over training time.}
    \label{fig:7}
\end{figure}

\begin{figure}[t]
    \centering
    \includegraphics[width=0.8\columnwidth]{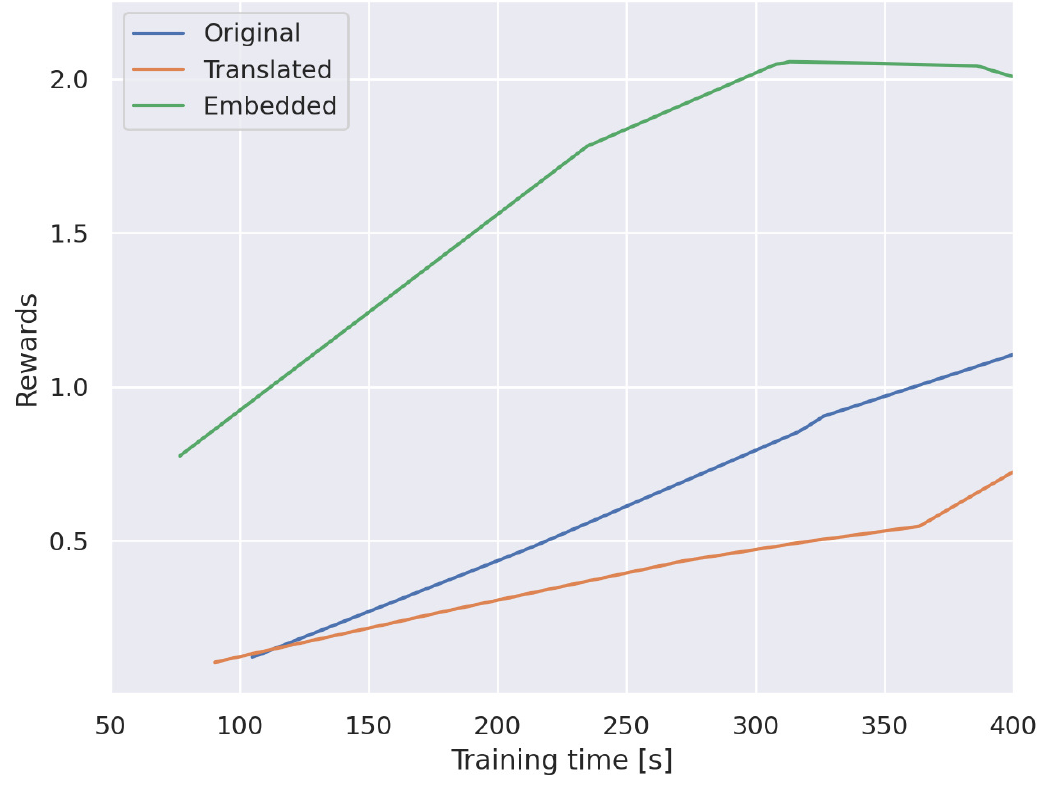}
    \caption{Task rewards relative to time}
    \label{fig:8}
\end{figure}

\subsection{Results on tissue manipulation learning}
This section presents the results of the experiment described in Section 3.4. First, the transition of loss during training is shown in Figure~\ref{fig:7}. Since the training time varies by condition, the results up to 400 seconds are smoothed using LOWESS \cite{cleveland1979robust}. The figure shows that Original and Translated have similar loss trends, with Translated exhibiting greater loss oscillations. In contrast, Embedded shows a much faster convergence of loss compared to the other two conditions, indicating an improvement in learning speed.

The evolution of task rewards across training time is shown in Figure~\ref{fig:8}, also smoothed for up to 400 seconds, similar to the loss transition. The figure shows that Embedded obtains higher rewards from the early stages of training compared to the other two conditions, with a significant increase in rewards over time. Comparing Original and Translated, there is no initial difference in rewards, but Original shows a greater increase in rewards during training.

\begin{figure}[t]
    \centering
    \includegraphics[width=0.7\columnwidth]{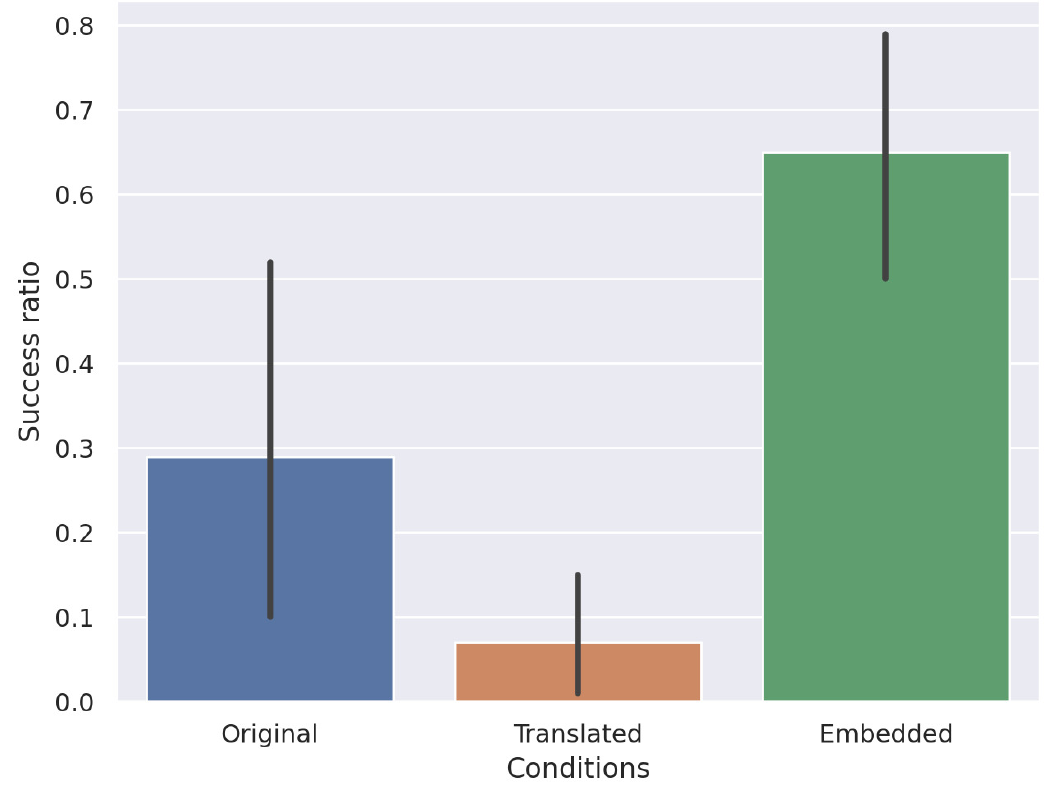}
    \caption{Average task success rate for the best-performing model saved during training.}
    \label{fig:9}
\end{figure}

\begin{figure}[t]
    \centering
    \includegraphics[width=0.7\columnwidth]{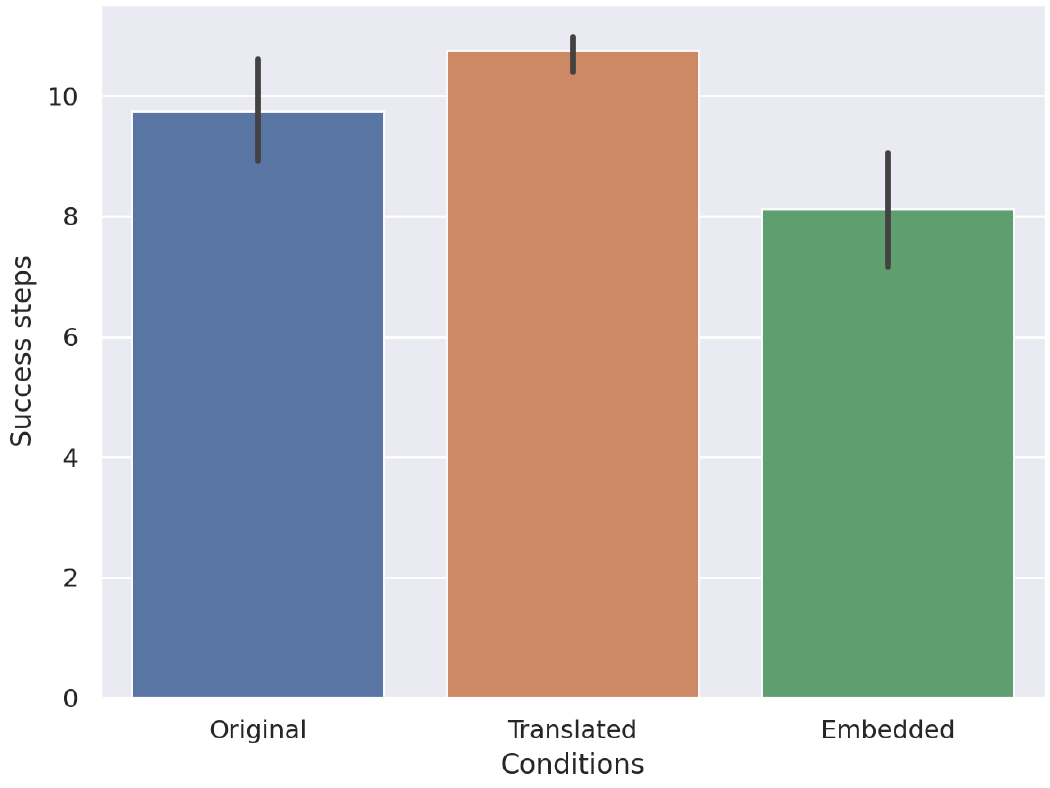}
    \caption{Average number of steps required for task success for the best-performing model saved during training.}
    \label{fig:10}
\end{figure}

Furthermore, the success rate of the task and the number of steps to success of the best-performing model saved during each training are averaged for each condition and shown in Figures~\ref{fig:9} and \ref{fig:10}, respectively. Both metrics show that Embedded outperforms Original and Translated, with Embedded achieving a task success rate of approximately 65\% for the triangulation task.

\begin{figure}[t]
    \centering
    \includegraphics[width=0.9\columnwidth]{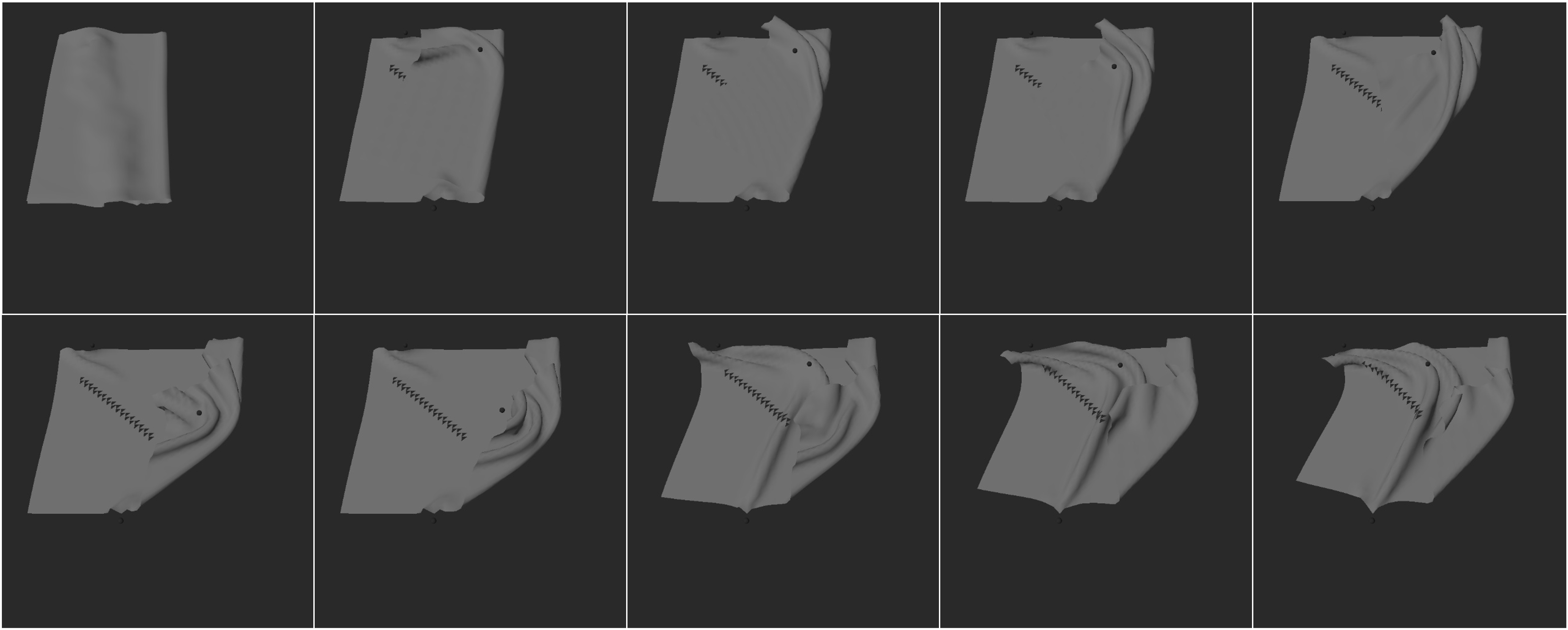}
    \caption{Simulator images during test episodes for a tissue triangulation task execution in the Embedded configuration.}
    \label{fig:11}
\end{figure}

\begin{figure}[t]
    \centering
    \includegraphics[width=0.95\columnwidth]{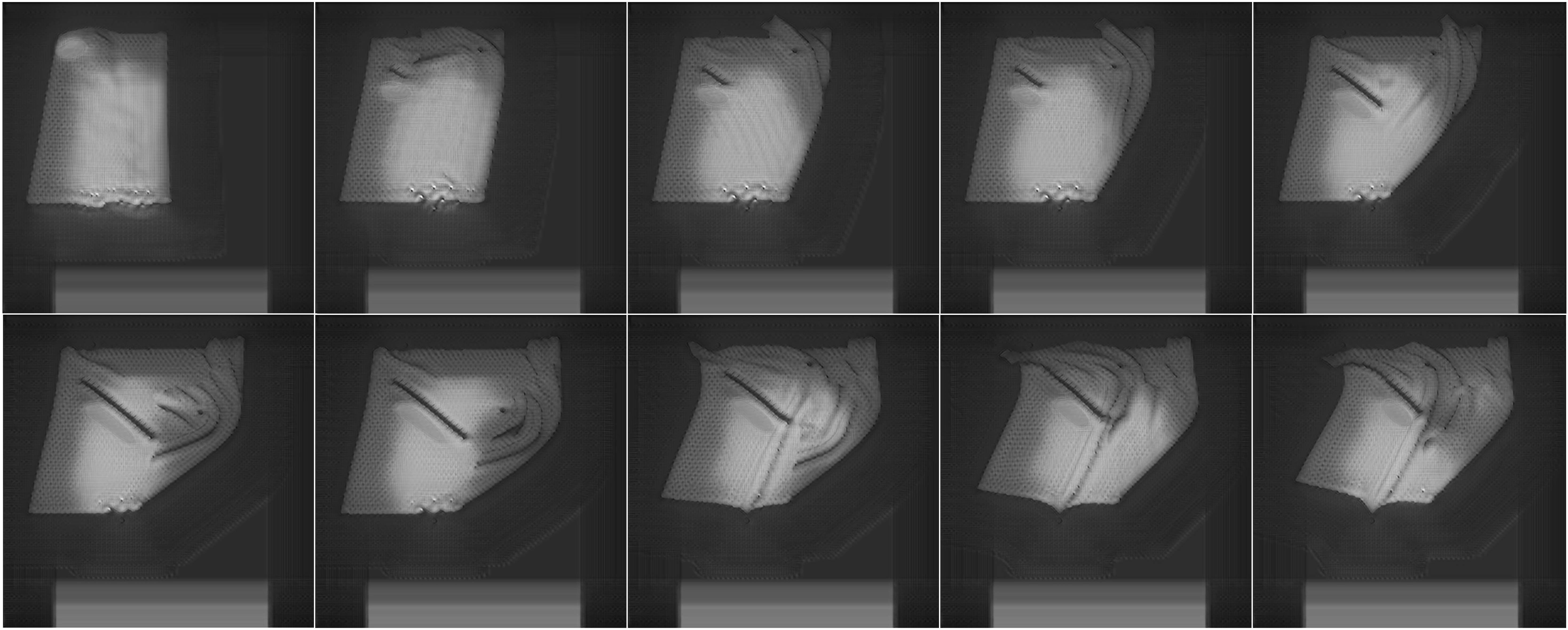}
    \caption{Translated images during test episodes for a tissue triangulation task execution in the Embedded configuration.}
    \label{fig:12}
\end{figure}

Figures~\ref{fig:11} and \ref{fig:12} show simulation images and translated images, respectively, during test episodes for the models trained with Embedded. The figures indicate that the image translation model successfully generates real-world images that maintain the tissue state in the simulation images, achieving SBML with real-world images.

\section{Conclusions}
Our results demonstrated that the proposed system significantly reduces the sim-to-real gap, with the embedded representation learning showing the most promising improvements in task success rates and training efficiency. The use of image translation models not only improved the visual realism of the simulated images, but also provided a robust foundation for training high-performance surgical models.

\addtolength{\textheight}{-5cm}   


\section*{ACKNOWLEDGMENT}
This work was supported in part by JST CREST Grant JPMJCR20D5, and in part by JSPS KAKENHI Grant 22K14221.




\end{document}